\documentclass[10pt,twocolumn,letterpaper]{article}

\usepackage{iccv}
\usepackage{times}
\usepackage{epsfig}
\usepackage{graphicx}
\usepackage{amsmath}
\usepackage{amssymb}

\usepackage{multirow}
\usepackage{float}
\usepackage{color}
\usepackage{xcolor}
\usepackage{colortbl}
\usepackage{url}
\usepackage[american]{babel}

\usepackage[pagebackref=true,breaklinks=true,letterpaper=true,colorlinks,bookmarks=false]{hyperref}

\iccvfinalcopy 

\ificcvfinal\fi

\newcommand{\pheadB}[1] {\vspace{1mm}\noindent\textbf{#1}}

\begin{document}
	
	\title{PP-YOLO: An Effective and Efficient Implementation of Object Detector}
	
	\author{
		Xiang Long, Kaipeng Deng, Guanzhong Wang, Yang Zhang, Qingqing Dang,  \\
		Yuan Gao, Hui Shen, Jianguo Ren, Shumin Han, Errui Ding, Shilei Wen\\
		{\tt\small
			\{longxiang, dengkaipeng, wangguanzhong, zhangyang57, dangqingqing,
		}\\
		{\tt\small
			gaoyuan18, shenhui08, v\_renjianguo, hanshumin, dingerrui, wenshilei\}@baidu.com
		} \\
		Baidu Inc.
	}
	
	\maketitle
	\begin{abstract}
		Object detection is one of the most important areas in computer vision, which plays a key role in various practical scenarios. Due to limitation of hardware, it is often necessary to sacrifice accuracy to ensure the infer speed of the detector in practice. Therefore, the balance between effectiveness and efficiency of object detector must be considered. The goal of this paper is to implement an object detector with relatively balanced effectiveness and efficiency that can be directly applied in actual application scenarios, rather than propose a novel detection model. Considering that YOLOv3 has been widely used in practice, we develop a new object detector based on YOLOv3. We mainly try to combine various existing tricks that almost not increase the number of model parameters and FLOPs, to achieve the goal of improving the accuracy of detector as much as possible while ensuring that the speed is almost unchanged. Since all experiments in this paper are conducted based on PaddlePaddle, we call it PP-YOLO. By combining multiple tricks, PP-YOLO can achieve a better balance between effectiveness (45.2\% mAP) and efficiency (72.9 FPS), surpassing the existing state-of-the-art detectors such as EfficientDet and YOLOv4.
		Source code is at \url{https://github.com/PaddlePaddle/PaddleDetection}.
	\end{abstract}

	\section{Introduction} \label{sec:intro}
	
	Object detection is an important yet challenging task. In the past few years, thanks to the advance of deep convolutional neural network\cite{krizhevsky2012imagenet,resnet}, object detectors have achieved remarkable performance\cite{faster-rcnn,fpn,yolov2,yolov3,yolov4,retinanet,SSD16,Fu2016dssd,freeanchor,cascade-rcnn,rfcn,GuidedAnchoring,tridentnet,GaussianYOLO2019,IoU-Net18,Mingxing2020EfficientDet}.
	
	\begin{figure}[t]
		\begin{center}
			\includegraphics[width=1.0\linewidth]{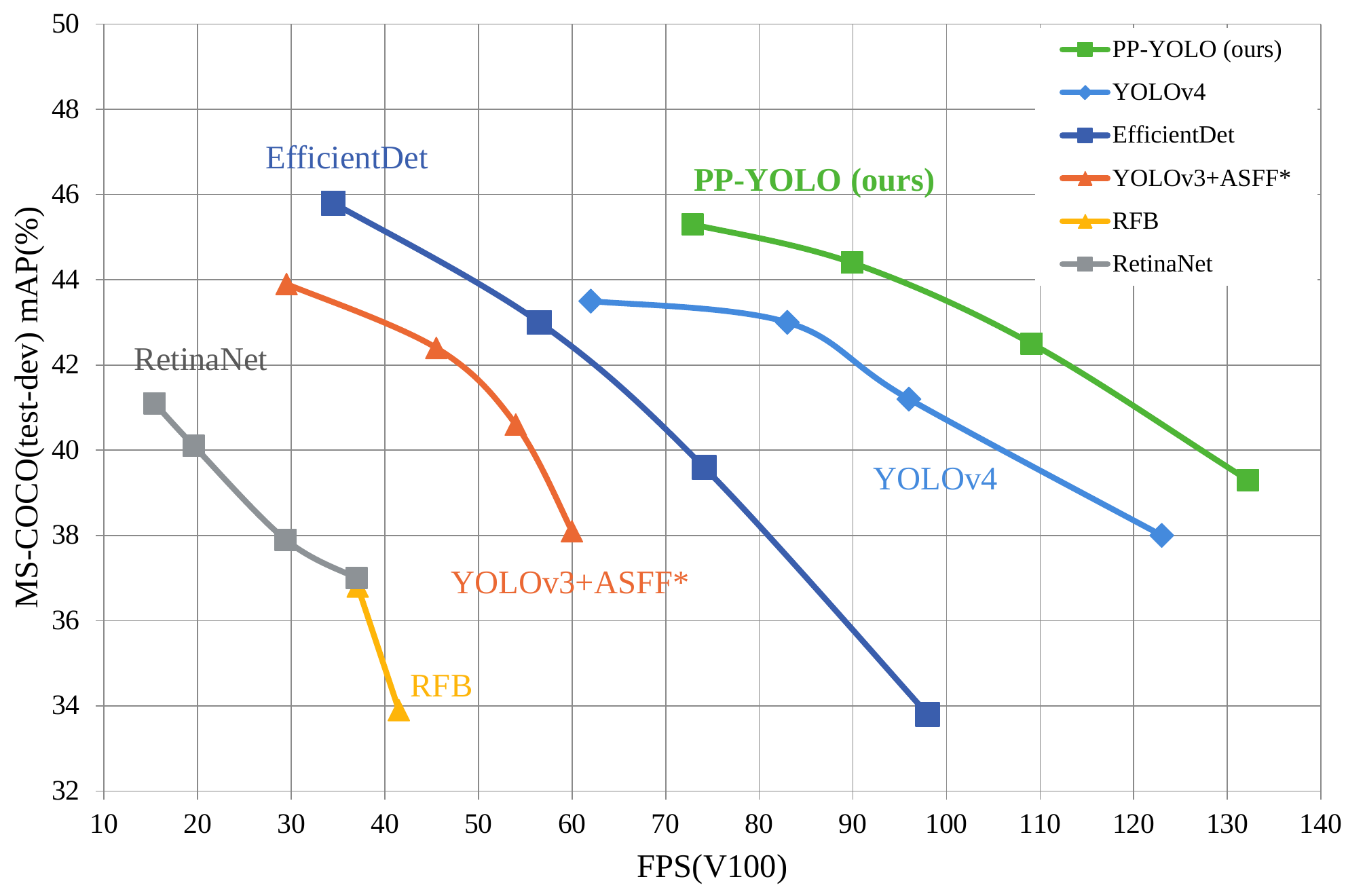}
		\end{center}
		\caption{Comparison of the proposed  PP-YOLO and other state-of-the-art object detectors.  PP-YOLO runs faster than YOLOv4 and improves mAP from 43.5\% to 45.2\%.}
		\label{fig:fps}
	\end{figure}
	
	In particular, one stage object detectors have a good balance between speed and accuracy, and have been widely used in practice\cite{ssd, retinanet, YOLO16, yolov2, yolov3, yolov4}. 
	YOLO series, including YOLOv1\cite{YOLO16}, YOLOv2\cite{yolov2}, YOLOv3\cite{yolov3}  and YOLOv4\cite{yolov4}, is one of the most famous series. Among them, the network structures of YOLO to YOLOv3 have relatively large changes. YOLOv4 considers various strategies such as bag of freebies and bag of specials on the basis of YOLOv3, which greatly improves the performance of the detector.
	This paper introduces an improved YOLOv3 model based on PaddlePaddle (PP-YOLO). A bunch of tricks that almost not increase the infer time are added to improve the overall performance of the model.
	
	Unlike YOLOv4, we did not explore different backbone networks and data augmentation methods, nor did we use NAS to search for hyperparameters.
	For the backbone, we directly use the most common ResNet\cite{resnet} as the backbone of PP-YOLO. For data augmentation, we directly used the most basic MixUp \cite{mixup}.
	One reason is that ResNet is used more wildly, such that various deep learning frameworks have deeply optimized for ResNet series, which will be more convenient in actual deployment and will have better infer speed in practical.
	Another reason is that the replacement of backbone and data augmentation are relatively independent factors, almost irrelevant to the tricks discussed in this paper. Since there are already a lot of works to study backbone network and to explore data augmentation, we do not repeat them in this paper.
	Searching for hyperparameters using NAS often consumes more computing power, so there is usually no condition to use NAS to perform a hyperparameter search in each new scenario. Therefore, we still use the manually set parameters following YOLOv3\cite{yolov3}.
	We believe that using a better backbone network, using more effective data augmentation method and using NAS to search for hyperparameters can further improve the performance of PP-YOLO.
	
	The focus of this paper is how to stack some effective tricks that hardly affect efficiency to get better performance. Many of these tricks cannot be directly applied to the network structure of YOLOv3, so small modification is required. Moreover, where to add tricks also needs careful consideration and experiment. This paper is not intended to introduce a novel object detecotor. It is more like a recipe, which tell you how to build a better detector step by step. We have found some tricks that are effective for the YOLOv3 detector, which can save developers' time of trial and error. The final PP-YOLO model improves the mAP on COCO from 43.5\% to 45.2\% at a speed faster than YOLOv4. The code and model is released in the PaddleDetection code-base (\url{https://github.com/PaddlePaddle/PaddleDetection}).
	
	\section{Related Work}
	
	Anchor-based methods are still the mainstream of object detection \cite{faster-rcnn,fpn,yolov2,yolov3,yolov4,retinanet,SSD16,Fu2016dssd,freeanchor,cascade-rcnn,rfcn,GuidedAnchoring,tridentnet,GaussianYOLO2019,IoU-Net18}, which evolved from early proposal based detectors, such as Fast R-CNN \cite{FastRCNN15}. 
	Their core idea is to introduce anchor boxes, which can be viewed as pre-defined proposals,  as a priori for bounding box regression. It mainly includes two branches: one-stage detectors and two-stage detectors\cite{Survey2019}. A large amount of one-stage detectors including YOLOv2\cite{yolov2}, YOLOv3\cite{yolov3}, YOLOv4\cite{yolov4}, RetinaNet \cite{retinanet}, RefineDet \cite{refinedet}, EfficentDet \cite{Mingxing2020EfficientDet}, FreeAnchor \cite{freeanchor}, and two-stage detectors including faster R-CNN \cite{faster-rcnn} FPN\cite{fpn}, Cascade R-CNN\cite{cascade-rcnn}, Trident-Net\cite{tridentnet} are proposed to promote the growth of state-of-the-art performance in object detection continuously.
	Besides,  anchor-free detectors have recently received more and more attention. In the past two years, a large number of new anchor-free methods have been proposed.
	The anchor-free method actually has a long history. Earlier works such as YOLOv1\cite{YOLO16}, DenseBox\cite{densebox} and UnitBox\cite{unitbox} can be considered as early anchor-free detectors.  They can be divided into two types. Anchor-point based detectors perform object bounding box regression based on anchor points instead of anchor boxes, including FSAF \cite{fsaf}, FCOS\cite{fcos}, FoveaBox\cite{foveabox}, SAPD\cite{sapd}. Keypoint based detectors reformulate the object detection as keypoints localization problem, including CornerNet\cite{cornernet}, CenterNet\cite{centernet}, ExtremeNet\cite{extremenet} and RepPoint\cite{reppoints}. Breaking the limitation imposed by hand-craft anchors, anchor-free methods show great potential for extreme object scales and aspect ratios \cite{mal}. The performance of some recently proposed anchor-free detectors can also compete with state-of-the-art anchor-based detectors. 
	
	YOLO series detectors \cite{YOLO16,yolov2,yolov3,yolov4}  have been widely used in practice, due to their excellent effectiveness and efficiency.
	Until the writing of this paper, it has developed to YOLOv4\cite{yolov4}. YOLOv4 discusses a large number of tricks including many ``bag of freebies'' which not increase the infer time, and several ``bag of specials'' that increase the inference cost by a small amount but can significantly improve the accuracy of object detection. YOLOv4 greatly improves the effectiveness and efficiency of the YOLOv3\cite{yolov3}. This paper is also developed based on YOLOv3 model and also explored a lot of tricks.  Unlike YOLOV4, we have not explored some widely studied parts such as data augmentation and backbone. Many tricks we discussed in this paper are different from YOLOV4 and the detailed implementation of tricks is also different.
	
	\begin{figure*}[t!]
		\centering
		\includegraphics[width=1.0\textwidth]{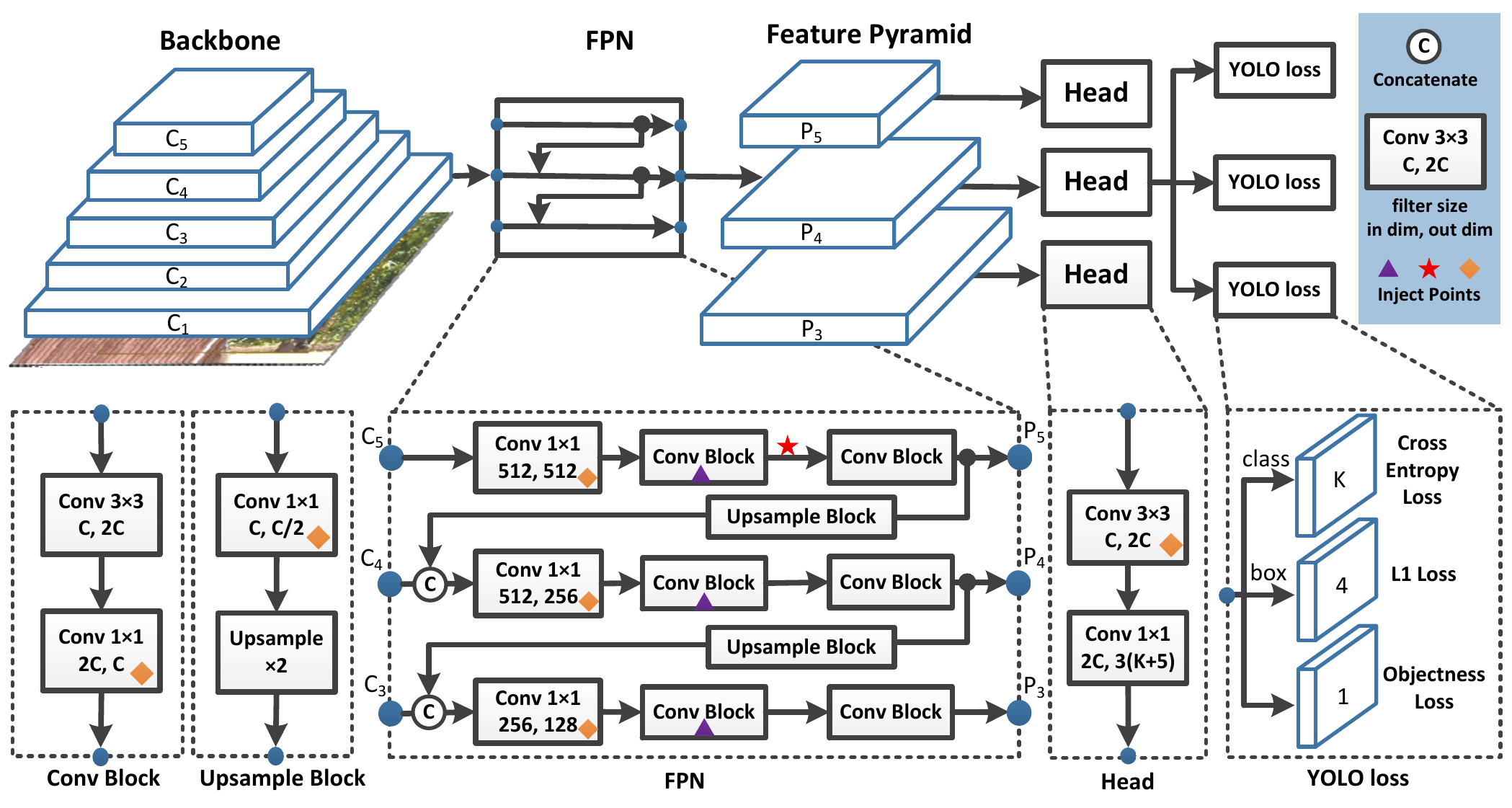}\\
		\caption{The network architecture of YOLOv3 and inject points for PP-YOLO. Activation layers are omitted for brevity. Details are described in Section \ref{sec:yolov3} and Section \ref{sec:tricks}.}
		\label{fig:yolov3}
	\end{figure*}

	\section{Method}
	An one-stage anchor-based detector is normally made up of a backbone network, a detection neck, which is typically a feature pyramid network (FPN), and a detection head for object classification and localization. They are also common components in most of the one-stage anchor-free detectors based on anchor-point. 
	We first revise the detail structure of YOLOv3 and introduce a modified version which replace the backbone to ResNet50-vd-dcn, which is used as the basic baseline in this paper.  Then we introduce a bunch of tricks which can improve the performance of YOLOv3 almost without losing efficiency.
	
	\subsection{Architecture} \label{sec:yolov3}
	
	\pheadB{Backbone} 
	The overall architecture of YOLOv3 is shown in Fig.~\ref{fig:yolov3}.
	In original YOLOv3\cite{yolov3}, DarkNet-53 is first applied to extract feature maps at different scales. Since ResNet\cite{resnet} has been widely used and and has been studied more extensively, there are more different variants for selection, and it has also been better optimized by deep learning frameworks.
	So, we replace the original backbone DarkNet-53 with ResNet50-vd in PP-YOLO. 
	Considering directly replace DarkNet-53 with ResNet50-vd will hurt the performance of YOLOv3 detector. We replace some convolutional layers in ResNet50-vd with deformable convolutional layers. The effectiveness of Deformable Convolutional Networks (DCN) has been verified in many detection models. DCN itself will not significantly increase the number of parameters and FLOPs in the model, but in practical application, too many DCN layers will greatly increase infer time. Therefore, in order to balance the efficiency and effectiveness, we only replace $3\times3$ convolution layers in the last stage with DCNs. We denote this modified backbone as ResNet50-vd-dcn, and the output of stage 3, 4 and 5 as $C_3, C_4, C_5$.
	
	\pheadB{Detection Neck}
	Then the FPN \cite{fpn} is used to build an feature pyramid with lateral connections between feature maps. Feature maps $C_3, C_4, C_5$ are  input to the FPN module. We denote the output feature maps of pyramid level $l$ as $P_l$, where  $l = 3, 4, 5$ in our experiments. The resolution of $P_l$ is $\frac{W}{2^l} \times \frac{H}{2^l}$ for an input image of size $W \times H$. The detail structure of FPN is shown in Fig.~\ref{fig:yolov3}.  
	
	\pheadB{Detection Head} 
	The detection head of YOLOv3 is very simple. It consists of two convolutional layers. A $3\times3$ convolutional followed by an $1\times1$ convolutional layer is adopt to get the final predictions. The output channel of each final prediction is $3(K+5)$, where $K$ is number of classes. Each position on each final prediction map has been associate with three different anchors. For each anchor, the first $K$ channels are the prediction of probability for $K$ classes. The following $4$ channels are the prediction for bounding box localization. The last channel is the prediction of objectness score. For classification and localization, cross entropy loss and L1 loss is adopt correspondingly. An objectness loss \cite{yolov3} is applied to supervise objectness score, which is used to identify whether is there an object or not. 
	
	\subsection{Selection of Tricks}  \label{sec:tricks}
	The various tricks we used in this paper are described in this section. These tricks are all already existing, which coming from different works \cite{dropblock,yolov4,iouloss,iouaware,solov2,coordconv,spp}. This paper does not propose an novel detection method, but just focuses on combining the existing tricks to implement an effective and efficient detector. Because many tricks cannot be applied to YOLOv3 directly, we need to adjust them according to the its structure.
	
	\pheadB{Larger Batch Size}
	Using a larger batch size can improve the stability of training and get better results. Here we change the training batch size from 64 to 192, and adjust the training schedule and learning rate accordingly. 
	
	\pheadB{EMA} 
	When training a model, it is often beneficial to maintain moving averages of the trained parameters. Evaluations that use averaged parameters sometimes produce significantly better results than the final trained values  \cite{Mingxing2020EfficientDet}.
	The Exponential Moving Average (EMA) compute the moving averages of trained parameters using exponential decay. For each parameter $W$, we maintain an shadow parameter
	\begin{equation}
	W_{EMA} = \lambda W_{EMA} + (1 - \lambda) W,
	\end{equation}
	where $\lambda$ is the decay.  We apply EMA with decay $\lambda$ of $0.9998$ and use the shadow parameter $W_{EMA}$ for evaluation.
	
	\pheadB{DropBlock} \cite{dropblock}
	DropBlock is a form of structured dropout, where units in a contiguous region of a feature map are dropped together. Different from the original paper, we only apply DropBlock to the FPN, since we find that adding DropBlock to the backbone will lead to a decrease of the performance.
	The detailed inject points of the DropBlock are marked by "triangles" in Figure \ref{fig:yolov3}.
	
	\pheadB{IoU Loss} \cite{iouloss}
	Bounding box regression is the crucial step in object detection. In YOLOv3, L1 loss is adopted for bounding box regression. It is not tailored to the mAP evaluation metric, which is strongly rely on Intersection over Union (IoU). IoU loss and other variations such as CIoU loss and GIoU loss\cite{diouloss,giouloss} have been proposed to address this problem. Different from YOLOv4, we do not replace the L1-loss with IoU loss directly, we add another branch to calculate IoU loss. We find that the improvements of various IoU loss are similar, so we choose the most basic IoU loss \cite{iouloss}. 
	
	\pheadB{IoU Aware} \cite{iouaware}
	In YOLOv3, the classification probability and objectness score is multiplied as the final detection confidence, which do not consider the localization accuracy. 
	To solve this problem, an IoU prediction branch is added to measure the accuracy of localization. During training, IoU aware loss is adopt to training the IoU prediction branch.
	During inference, the predicted IoU is multiplied by the classification probability and objectiveness score to compute the final detection confidence, which is more correlated with the localization accuracy. The final detection confidence is then used as the input of the subsequent NMS. IoU aware branch will add additional computational cost. However, only 0.01\% number of parameters and 0.0001\% FLOPs are added, which can be almost ignored.
	
	\pheadB{Grid Sensitive}\cite{yolov4}
	Grid Sensitive is an effective trick introduced by YOLOv4. 
	When we decode the coordinate of the bounding box center $x$ and $y$, in original YOLOv3, we can get them by
	\begin{align}
	x = s \cdot ( g_x + \sigma(p_x) ),\\
	y = s \cdot ( g_y + \sigma(p_y) ),
	\end{align}
	where $\sigma$ is the sigmoid function, $g_x$ and $g_y$ are integers and $s$ is a scale factor. Obviously, $x$ and $y$ cannot be exactly equal to $s \cdot g_x$ or $s \cdot (g_x+1)$.
	This makes it difficult to predict the centres of bounding boxes that just located on the grid boundary. We can address this problem, by change the equation to 
	\begin{align}
	x = s \cdot ( g_x + \alpha \cdot  \sigma(p_x) - (\alpha-1)/2 ),\\
	y = s \cdot ( g_y + \alpha \cdot  \sigma(p_y) - (\alpha-1)/2 ),
	\end{align}
	where $\alpha$ is set to $1.05$ in this paper. This makes it easier for the model to predict bounding box center exactly located on the grid boundary. The FLOPs added by Grid Sensitive is really small, and can be totally ignored.
	
	\pheadB{Matrix NMS} \cite{solov2}
	Matrix NMS is motivated by Soft-NMS, which decays the other detection scores as amonotonic decreasing function of their overlaps. However, such process is sequential like traditional Greedy NMS and could not be implemented in parallel. Matrix NMS views this process from another perspective and implement it in a parallel manner.
	Therefore, the Matrix NMS is faster than traditional NMS, which will not bring any loss of efficiency.
	
	\pheadB{CoordConv} \cite{coordconv}
	CoordConv, which works by giving convolution access to its own input coordinates through the use of extra coordinate channels.  CoordConv allows networks to learn either complete translation invariance or varying degrees of translation dependence. Considering that CoordConv will add two inputs channels to the convolution layer, some parameters and FLOPs will be added. In order to reduce the loss of efficiency as much as possible, we do not change convolutional layers in backbone, and only replace the 1x1 convolution layer in FPN and the first convolution layer in detection head with CoordConv.
	The detailed inject points of the CoordConv are marked by "diamonds" in Figure \ref{fig:yolov3}.
	
	\begin{table*}[t!]
		\centering
		\begin{tabular}{l|l|l|l|l|l|l}
			\hline
			& \textbf{Methods} & \textbf{mAP(\%)} & \textbf{Parameters} & \textbf{GFLOPs} & 
			\textbf{infer time} & \textbf{FPS} \\ 
			\hline
			\hline
			A & Darknet53 YOLOv3 & 38.9 &59.13 M & 65.52 & 17.2 ms & 58.2  \\ 
			\hline
			\hline
			B & ResNet50-vd-dcn YOLOv3 & 39.1 &43.89 M & 44.71 & 12.6 ms& 79.2 \\ 
			C & B + LB + EMA + DropBlock & 41.4 &43.89 M  & 44.71& 12.6 ms& 79.2\\ 
			D & C +  IoU Loss & 41.9 &43.89 M & 44.71 & 12.6 ms& 79.2 \\ 
			E & D + Iou Aware & 42.5 &43.90 M & 44.71 & 13.3 ms& 74.9\\ 
			F & E + Grid Sensitive& 42.8 &43.90 M & 44.71 & 13.4 ms& 74.8 \\ 
			G & F + Matrix NMS & 43.5 &43.90 M & 44.71& 13.4 ms& 74.8  \\ 
			H & G + CoordConv  & 44.0 &43.93 M & 44.76 & 13.5ms& 74.1 \\ 
			I & H + SPP & 44.3 &44.93 M & 45.12 & 13.7 ms& 72.9 \\
			J & I + Better ImageNet Pretrain  & 44.6 &44.93 M & 45.12 & 13.7 ms& 72.9  \\ 
			\hline
		\end{tabular}
		\caption{The ablation study of tricks on the MS-COCO minival split.}
		\label{tab1}
	\end{table*}
	
	\pheadB{SPP} \cite{spp}
	The Spatial Pyramid Pooling (SPP) is first proposed by He et al\cite{spp}. SPP integrates SPM into CNN and use max-pooling operation instead of bag-of-word operation. 
	YOLOv4 apply SPP module by concatenating max-pooling outputs with kernel size $k \times k$, where $k = \{1, 5, 9, 13\}$, and stride equals to 1. Under this design, a relatively large $k \times k$ max-pooling effectively increase the receptive field of backbone feature. In detail, the SPP only applied on the top feature map as shown in Figure \ref{fig:yolov3} with "star" mark. No parameter are introduced by SPP itself, but the number of input channel of the following convolutional layer will increase. So around 2\% additional papameters and 1\% extra FLOPs are introduced.
	
	\pheadB{Better Pretrain Model}
	Using a pretrain model with higher classification accuracy on ImageNet may result in better detection performance. Here we use the distilled ResNet50-vd model as the pretrain model \cite{pretrain} . This obviously does not affect the efficiency of the detector.

	\section{Experiment}
	
	In this section, we present the effectiveness of different tricks.
	Experiments were carried out on the bounding box detection track of the COCO dataset \cite{coco}. 
	Following the common practice \cite{yolov3,Mingxing2020EfficientDet,yolov4}, we use \texttt{trainval35k} split for training, which contains $\sim \! \! 118k$ images, \texttt{minival} split ($5k$) for validation and ablation study, and \texttt{test-dev} split($\sim \! \! 20k$) for testing. 
	
	\subsection{Implementation Details} \label{sec:detail}
	
	We use ResNet50-vd-dcn\cite{resnet} as the backbone networks unless specified. The architecture of FPN and head in our basic models is completely the same as YOLOv3\cite{yolov3}. The details have been presented in section \ref{sec:yolov3}.
	We initialize our detectors following common practice. Specifically, our backbone networks are initialized with the weights pre-trained
	on ImageNet\cite{imagenet}. For the FPN and detection heads, we initialize them randomly as same as in YOLOv3\cite{yolov3}.
	For the baseline model (A, B), The training schedule is as same as YOLOv3. Under larger batch size setting, the entire network is trained with stochastic gradient descent (SGD) for 250K iterations with the initial learning rate being 0.01 and a minibatch of 192 images distributed on 8 GPUs. The learning rate is divided by 10 at iteration 150K and 200K, respectively. Weight decay is set as 0.0005, and momentum is set as 0.9. 
	Multi-scale training from 320 to 608 pixels is applied. MixUp\cite{mixup} is adopted for data augmentation.

	\begin{table*}[h]
		\centering
		\resizebox{1.0\textwidth}{!}{
			\begin{tabular}{l|l|c|cc|cccccc}
				\hline
				\multirow{2}{*}{\textbf{Method}} & \multirow{2}{*}{\textbf{Backbone}} & \multirow{2}{*}{\textbf{Size}} &\multicolumn{2}{c|}{\textbf{FPS (V100)}} &
				\multirow{2}{*}{\textbf{AP}} & \multirow{2}{*}{\textbf{AP$_{50}$}} & \multirow{2}{*}{\textbf{AP$_{75}$}} & \multirow{2}{*}{\textbf{AP$_S$}} & \multirow{2}{*}{\textbf{AP$_M$}} & \multirow{2}{*}{\textbf{AP$_L$}}\\	
				\cline{4-5} 
				& & & \textbf{w/o TRT} & \textbf{with TRT} & & & & & &\\	
				\hline
				\hline
				RetinaNet \cite{retinanet} & ResNet-50 & 640 &  37  & -  & 37.0\% & - & - & - & - & - \\
				RetinaNet \cite{retinanet} & ResNet-101 & 640  & 29.4 & -  & 37.9\% & - & - & - & - & - \\
				RetinaNet \cite{retinanet} & ResNet-50 & 1024 & 19.6 & -  & 40.1\% & - & - & - & - & - \\
				RetinaNet \cite{retinanet} & ResNet-101 & 1024  & 15.4  & - & 41.1\% & - & - & - & - & - \\
				\hline
				EfficientDet-D0 \cite{Mingxing2020EfficientDet} & Efficient-B0  & 512 &  98.0$^+$ & - & 33.8\% & 52.2\% & 35.8\% & 12.0\% & 38.3\% & 51.2\% \\
				EfficientDet-D1 \cite{Mingxing2020EfficientDet} & Efficient-B1 & 640 &  74.1$^+$  &- & 39.6\% & 58.6\% & 42.3\% & 17.9\% & 44.3\% & 56.0\% \\
				EfficientDet-D2 \cite{Mingxing2020EfficientDet} & Efficient-B2 & 768 & 56.5$^+$  &- & 43.0\% & 62.3\% & 46.2\% & 22.5\% & 47.0\% & 58.4\% \\
				EfficientDet-D2 \cite{Mingxing2020EfficientDet} & Efficient-B3 & 896 & 34.5$^+$  &- & 45.8\% & 65.0\% & 49.3\% & 26.6\% & 49.4\% & 59.8\% \\
				\hline
				RFBNet\cite{Chao2019HarDNet} & HarDNet68 & 512 &  41.5   & - &33.9\% & 54.3\% & 36.2\% & 14.7\% & 36.6\% & 50.5\% \\
				RFBNet\cite{Chao2019HarDNet} & HarDNet85 & 512 &  37.1   & - &36.8\% & 57.1\% & 39.5\% & 16.9\% & 40.5\% & 52.9\% \\
				\hline
				YOLOv3 + ASFF* \cite{Liu2019Learning} & Darknet-53 & 320 &  60   &- & 38.1\% & 57.4\% & 42.1\% & 16.1\% & 41.6\% & 53.6\% \\
				YOLOv3 + ASFF* \cite{Liu2019Learning}& Darknet-53 & 416 &  54   & - &40.6\% & 60.6\% & 45.1\% & 20.3\% & 44.2\% & 54.1\% \\
				YOLOv3 + ASFF* \cite{Liu2019Learning}& Darknet-53 & 608 &  45.5   & - &42.4\% & 63.0\% & 47.4\% & 25.5\% & 45.7\% & 52.3\% \\
				YOLOv3 + ASFF* \cite{Liu2019Learning} & Darknet-53 & 800 & 29.4   &- & 43.9\% & 64.1\% & 49.2\% & 27.0\% & 46.6\% & 53.4\% \\
				\hline
				YOLOv4 \cite{yolov4} & CSPDarknet-53 & 416 & 96  &164.0$^*$ & 41.2\% & 62.8\% & 44.3\% & 20.4\% & 44.4\% & 56.0\% \\
				YOLOv4  \cite{yolov4} & CSPDarknet-53 & 512 & 83  &138.4$^*$  & 43.0\% & 64.9\% & 46.5\% & 24.3\% & 46.1\% & 55.2\% \\
				YOLOv4  \cite{yolov4} & CSPDarknet-53 & 608  & 62  &105.5$^*$  & 43.5\% & 65.7\% & 47.3\% & 26.7\% & 46.7\% & 53.3\% \\
				\hline
				\hline
				PP-YOLO & ResNet50-vd-dcn & 320 & 132.2 &  242.2 & 39.3\% & 59.3\% & 42.7\% & 16.7\% & 41.4\% & 57.8\% \\
				PP-YOLO & ResNet50-vd-dcn & 416 & 109.6 & 215.4 & 42.5\% & 62.8\% & 46.5\% & 21.2\% & 45.2\% & 58.2\% \\
				PP-YOLO & ResNet50-vd-dcn & 512 & 89.9 & 188.4 & 44.4\%  & 64.6\% & 48.8\% & 24.4\% & 47.1\% & 58.2\% \\
				PP-YOLO& ResNet50-vd-dcn & 608 & 72.9 &  155.6 & 45.2\% & 65.2\% & 49.9\% & 26.3\% & 47.8\% & 57.2\% \\
				\hline
			\end{tabular}
		}
		\caption{Comparison of the speed and accuracy of different object detectors on the MS-COCO (test-dev 2017). We compare the results with batch size = 1, without tensorRT (w/o TRT) or with tensorRT(with TRT).
			Results marked by "+" are updated results from the corresponding official code base, which are higher than the results in original paper.
			Results marked by "*" are test in our environment using official code and model, which are slightly higher than results reported in official code-base.
		}
		\label{tab2}
	\end{table*}
	
	\subsection{Ablation Study}
	
	In this section, we present the effectiveness of each module in an incremental manner. The reason is that each trick is not completely independent. Some tricks are effective when applied alone, but they are not effective when combined together. Since there are too many combinations of various tricks, it is difficult to conduct a comprehensive analysis.
	Therefore, we show how to improve the performance of the object detector step by step in the order of our exploration and discovering the effectiveness of tricks. Results are shown in Table \ref{tab1}, where infer time and FPS do not consider the influence of NMS following YOLOv4\cite{yolov4}.
	
	\pheadB{A $\rightarrow$ B} 
	First of all, we try to build a basic version of PP-YOLO. Because the ResNet\cite{resnet} series is more widely used, we first replace the original YOLOv3 backbone Darknet53 with ResNet50-vd. However, we found that it will cause a significant decrease in mAP. Considering that the number of parameters and FLOPs of ResNet50-vd are much smaller than those of Darknet53, we replace the $3\times3$ convolutional layer in the last stage of ResNet with deformable convolution layer\cite{dcn}. In this way, we get a basic PP-YOLO model (B) with a mAP of 39.1\%, which is slightly higher than the original YOLOv3 (A), but its parameters, FLOPs and infer time are much smaller than the original YOLOv3 model.
	
	\pheadB{B $\rightarrow$ C} 
	We first try to optimize the training strategy. We use a larger batch size and EMA to improve the stability of the model, and also apply DropBlock to prevent the model from overfitting. After using these strategies, the mAP of model (C) increases to 41.4\% without any loss of efficiency.
	
	\pheadB{C $\rightarrow$ F} 
	Next, we consider modifying the YOLO loss to improve the effectiveness of the model, because modifying the loss generally only has an impact on the training process, and will not or rarely affect the infer time.
	We add IoU Loss (D), IoU Aware (E) and Grid Sensitive (F) modules, and increase the mAP by 0.5\%, 0.6\% and 0.3\% respectively. Among them, IoU loss will not affect the number of parameters and the infer time at all. IoU Aware and Grid Sensitive will increase the post-processing time by 0.7ms and 0.1ms, since the current implementation is not efficient enough, which can be greatly reduced by merging them as a single OP in PaddlePaddle in the future. On the whole, we have increased the mAP of PP-YOLO from 41.4\% to 42.8\%.
	
	\pheadB{F $\rightarrow$ G} 
	Post-processing is also a place where we can improve the performance. We use Matrix NMS (G) to replace traditional greedy NMS. We can see that the mAP has improved by 0.6\%. Since the infer time in Table \ref{tab1} does not consider NMS, so the influence is not shown here. In fact, the overall infer time is decreased since the efficiency of MatrixNMS is higher than traditional NMS.
	
	\pheadB{G $\rightarrow$ I} 
	It has become difficult to continue to improve mAP without increasing the number of parameters and FLOPs. So we considered two methods that only increase a few parameters and FLOPs but can bring effective improvements, CoordConv (H) and SPP (I).
	CoordConv will cause the input channel of convolutional layers increase by 2, the number of parameters increases by 0.03M, and FLOPs increases by 0.05G, which is very small compared to the whole model. It can bring an improvement of 0.5\% mAP. 
	SPP itself does not increase the parameters, but it will increase the input channel of the convolutional layer just following it, resulting in an increase of the parameters by 1M and FLOPs by 0.36G. It can improve the mAP of PP-YOLO by 0.3\% further. After adding these two modules, the infer time has increased by 0.3ms.
	
	\pheadB{I $\rightarrow$ J} 
	Replacing the pre-trained model is a very common approach. However, the accuracy of pretrained classification model is higher does not mean that the final detection model is more effective, and the degree of improvement will be affected by the tricks we used. So we consider it at the end.
	For fair comparisons, we still use ImageNet for pre-training. We use a distilled ResNet50-vd model for backbone initialization. The mAP of PP-YOLO can be further improved by 0.3\%. In fact, using other detection datasets for pre-training can greatly improve the performance of the model, but this is beyond the scope of this paper.
	
	\subsection{Comparison with Other State-of-the-Art Detectors}	
	
	Comparison of the results on MS-COCO test split with other state-of-the-art object detectors are shown in Figure \ref{fig:fps} and Table \ref{tab2}. The FPS results of PP-YOLO and other methods are all tested on V100 with batch size = 1. 
	We considered two different test conditions, without tensorRT (w/o TRT) and with tensorRT (with TRT). The test methods are consistent with YOLOv4\cite{yolov4}. 
	Results marked by "+" are updated results from the corresponding official code-base, which are higher than the results in original paper, 
	Results marked by "*" are test in our environment using official code and model, which are slightly higher than results reported in official code-base.
	
	Compared with other state-of-the-art methods, our PP-YOLO has certain advantages in speed and accuracy. For example, compared with YOLOv4, our PPYOLO can increased the mAP on COCO from 43.5\% to 45.2\% with FPS improved from 62 to 72.9. It is worth noticing that tensorRT accelerates the PP-YOLO model more obviously. The relative improvement of PP-YOLO (around 100\%) is larger than YOLOv4(around 70\%). We speculate that it is mainly because tensorRT optimizes for ResNet model better than Darknet.
	
	In addition, we can get a series of PP-YOLO results by changing the input size of the image. Here we also show the results for 320, 416, 512 and 608 input sizes. Figure \ref{fig:fps} shows that PP-YOLO results have advantages in the balance of speed and accuracy compared with other detectors.
	
	\section{Conclusions}
	
	This paper introduce a new implementation of object detector based on PaddlePaddle, called PP-YOLO.  PP-YOLO is faster (FPS) and more accurate(COCO mAP) than other state-of-the-art detectors, such as EfficientDet and YOLOv4. In this paper, we explore a lot of tricks and show how to combine these tricks on the YOLOv3 detector and demonstrate their effectiveness. We hope this paper can help developers and researchers save exploration time and get better performance in practical applications.
	
	\bibliographystyle{ieee}
	\bibliography{references}
\end{document}